\documentclass[conference]{IEEEtran}
\IEEEoverridecommandlockouts
\usepackage{cite}
\usepackage{amsmath,amssymb,amsfonts}
\usepackage{algorithmic}
\usepackage{graphicx}
\usepackage{threeparttable}
\usepackage{textcomp}
\usepackage{xcolor}
\usepackage{multirow}
\usepackage{booktabs}
\usepackage{subfigure}
\def\BibTeX{{\rm B\kern-.05em{\sc i\kern-.025em b}\kern-.08em
    T\kern-.1667em\lower.7ex\hbox{E}\kern-.125emX}}

\begin{document}

\title{Pathological Prior-Guided Multiple Instance Learning For Mitigating Catastrophic Forgetting in Breast Cancer Whole Slide Image Classification \\
\thanks{*: Corresponding Authors. \\ This work was supported in part by the National Key Research and Development Program of China (2023YFC2705700), NSFC 62222112, and 62176186, the Innovative Research Group Project of Hubei Province under Grants (2024AFA017), the Cross-disciplinary Innovation Talent Project of Renmin Hospital of Wuhan University (JCRCZN-2022-015).}}

\author{

\IEEEauthorblockN{Weixi Zheng}
\IEEEauthorblockA{\textit{School of Computer Science} \\
\textit{Wuhan University}\\
Wuhan, China \\
acnicotine@whu.edu.cn}
\and
\IEEEauthorblockN{Aoling Huang}
\IEEEauthorblockA{\textit{Department of Pathology} \\
\textit{Renmin Hospital of Wuhan University}\\
Wuhan, China \\
huangaoling@whu.edu.cn}

\and
\IEEEauthorblockN{Jingping Yuan}
\IEEEauthorblockA{\textit{Department of Pathology} \\
\textit{Renmin Hospital of Wuhan University}\\
Wuhan, China \\
yuanjingping@whu.edu.cn}

\and
\IEEEauthorblockN{Haoyu Zhao}
\IEEEauthorblockA{\textit{School of Computer Science} \\
\textit{Wuhan University}\\
Wuhan, China \\
haoyu.zhao@whu.edu.cn}
\and
\IEEEauthorblockN{Zhou Zhao}
\IEEEauthorblockA{\textit{School of Computer Science} \\
\textit{Central China Normal University}\\
Wuhan, China \\
zhaozhou@ccnu.edu.cn}
\and
\IEEEauthorblockN{Yongchao Xu\textsuperscript{*}}
\IEEEauthorblockA{\textit{School of Computer Science} \\
\textit{Wuhan University}\\
Wuhan, China \\
yongchao.xu@whu.edu.cn}

\and
\IEEEauthorblockN{Thierry G\'eraud}
\IEEEauthorblockA{
\textit{EPITA Research Laboratory}\\
Le Kremlin-Bic\^etre, France \\
thierry.geraud@epita.fr}

}

\maketitle

\begin{abstract}
In histopathology, intelligent diagnosis of Whole Slide Images (WSIs) is essential for automating and objectifying diagnoses, reducing the workload of pathologists. However, diagnostic models often face the challenge of forgetting previously learned data during incremental training on datasets from different sources. To address this issue, we propose a new framework PaGMIL to mitigate catastrophic forgetting in breast cancer WSI classification. Our framework introduces two key components into the common MIL model architecture. First, it leverages microscopic pathological prior to select more accurate and diverse representative patches for MIL. Secondly, it trains separate classification heads for each task and uses macroscopic pathological prior knowledge, treating the thumbnail as a prompt guide (PG) to select the appropriate classification head. We evaluate the continual learning performance of PaGMIL across several public breast cancer datasets. PaGMIL achieves a better balance between the performance of the current task and the retention of previous tasks, outperforming other continual learning methods. Our code will be open-sourced upon acceptance.
\end{abstract}


\section{Introduction}
WSIs provide crucial histopathological information for breast cancer diagnosis. Over 30 million breast cancer slides are analyzed annually, underscoring the need for intelligent WSI diagnosis. Deep learning advancements offer potential for automated diagnosis, prognosis, and treatment, reducing pathologists' workload \cite{cornish2012whole,shao2021transmil,tanizaki2016report}. However, the large size of WSIs and costly pixel-level annotations present challenges for deep learning models \cite{lucapturing,huang2022deep,javed2020multiplex}. To address this, CLAM\cite{lu2021data} has been proposed as an effective multiple instance learning (MIL) approach for WSI analysis. It utilizes slide-level labels to divide WSIs into patches, analyzes them, and aggregates results for slide-level predictions.
% ==============================================
% ==============================================

\begin{figure}[t]
  \centering
  \includegraphics[width=1\linewidth]{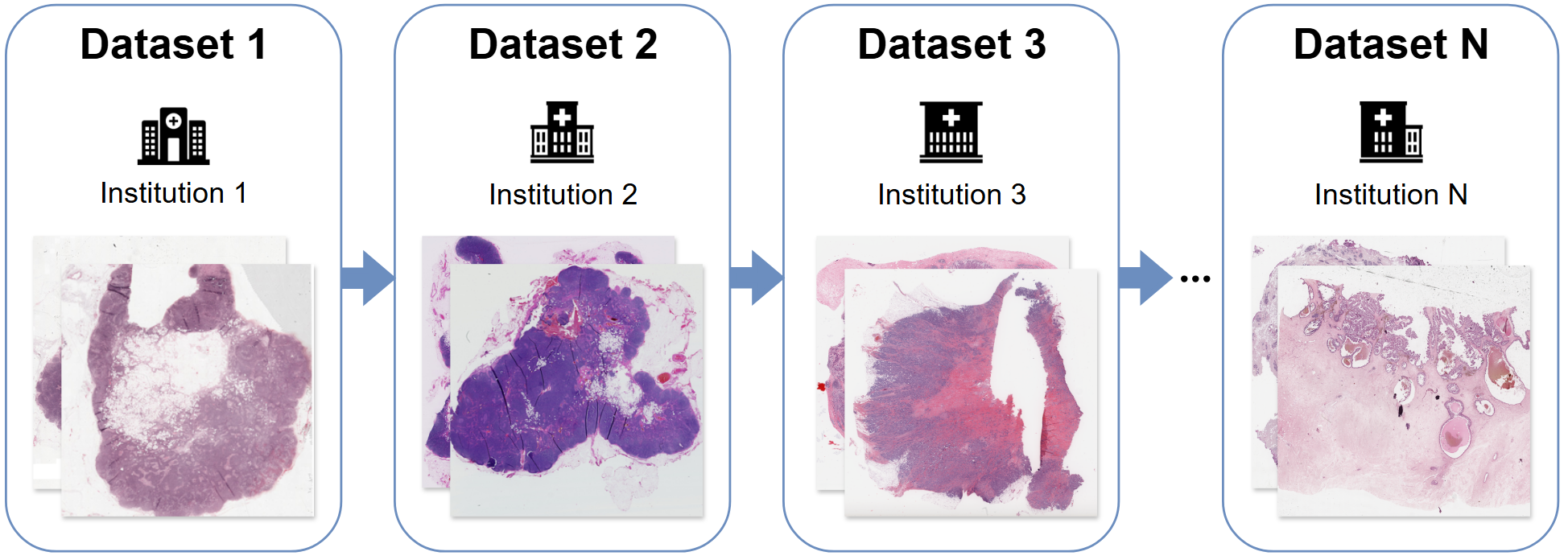}
\caption{Illustration  of dynamic input of WSI data. The actual model update involves different medical institutions sending in some data for training. Significant differences in data from different batches arise due to variations in staining methods or equipment used across different medical institutions.}
   \label{fig:dataset}
\end{figure}

% ==============================================
% ==============================================

Despite encouraging results, existing methods \cite{xu2024whole,chen2024towards,chen2022scaling,zheng2022kernel,guan2022node,shao2021transmil} typically use static models trained and tested on fixed datasets. However, WSI imaging is dynamic, influenced by variations in equipment and staining \cite{shen2022federated}, which limits model performance across different environments. In practical applications, models encounter pathology images from various medical institutions (see Fig. \ref{fig:dataset}), requiring adaptability to maintain robustness \cite{lee2020clinical,derakhshani2022lifelonger,kaustaban2022characterizing,van2021deep,perkonigg2021dynamic}. Fine-tuning pre-trained models on new datasets is one solution, but it often causes catastrophic forgetting, where models overfit to new data and lose knowledge from previous datasets \cite{boschini2022transfer,de2021continual,lesort2020continual}.

% ==============================================
% ==============================================

\begin{figure*}[htbp]
    \centering
    \includegraphics[width=1.0\linewidth]{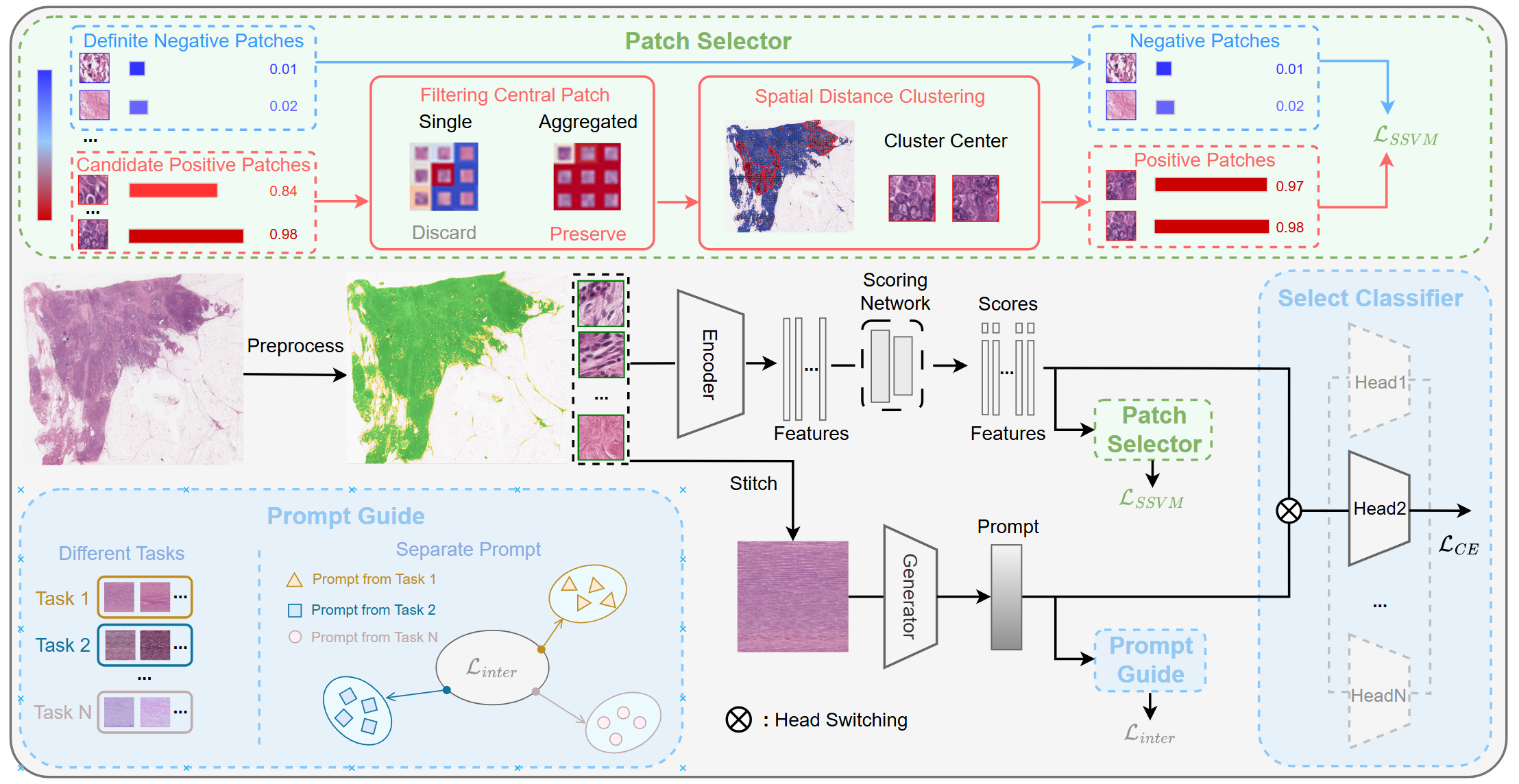}
    \caption{ Overview of our proposed PaGMIL method. Based on the MIL pipeline, we add two new modules. The Patch Selector module uses microscopic pathology prior to select more accurate representative patches, and the PG module uses macroscopic pathology priors to select the correct classification head.}
    \label{fig:pipeline}
\end{figure*}

% ==============================================
% ==============================================

Continual learning (CL) \cite{lopez2017gradient,de2021continual} aims to mitigate catastrophic forgetting by allowing models to adapt to new tasks while retaining knowledge from previous ones, enhancing robustness and accommodating data growth. CL has shown promise on natural images and includes approaches like parameter regularization \cite{kirkpatrick2017overcoming,rebuffi2017icarl,kurle2019continual}, knowledge distillation \cite{li2017learning,bonato2024mind,zajkac2023prediction}, dynamic network structure \cite{liang2024inflora,pham2021dualnet,douillard2022dytox}, and data replay \cite{li2024towards,kim2024sddgr,buzzega2020dark}. However, applying these CL methods to WSI images is challenging due to their large size and the presence of background noise, making data replay, regularization, and distillation less effective. In recent years, a data replay method\cite{huang2023conslide} is designed specifically for the characteristics of WSIs, achieving promising results.

In this paper, we propose PaGMIL, a novel MIL continual learning network that leverages both micro and macro pathological priors to guide the learning of new data and mitigate catastrophic forgetting. At the micro level, we introduce positional information by recording the coordinates of each patch. By incorporating prior knowledge about cancer clustering, it is possible to rule out misjudgments caused by staining bias or noise. At the same time, we select patches with large spatial distances to capture diverse tissue environments. This encourages the attention network to learn generalizable knowledge to mitigate forgetting. At the macro level, we introduce global information by converting the WSI into thumbnail and feeding it into a Generator to generate prompt. This prompt serves as global information to indicate the current task during classification and helps select the corresponding classification head.

In summary, the main contributions are threefold: 1) We introduce positional information to select more representative patches for continual learning. 2) We use global information to select the correct classification head. 3) We evaluate PaGMIL on four public breast cancer WSI datasets (\textit{TCGA}, \textit{CAMELYON16}, \textit{BRACS}, and \textit{BACH}) and demonstrate superior balance between the performance of the current task and the retention of previous tasks.

\section{Method}
\subsection{Overview}
Our proposed PaGMIL method, shown in Fig. \ref{fig:pipeline}, follows the general MIL approach of CLAM \cite{lu2021data}, with the addition of two modules: Patch Selector (PS) and Prompt Guide (PG). The PS module utilizes microscopic pathological priors to process ranked patches. It defines low-scoring patches as negative patches and identifies high-scoring patches as candidate positive patches for further screening. The PS module filters out isolated high-score patches, then clusters the remaining high-score patches into B categories based on their positions, and selects B cluster centers as positive patches. The PG module leverages macroscopic pathological priors to convert the WSI into a thumbnail, which is then processed to generate prompts. Prompts for the same task are positioned closer, while those for different tasks are spaced further apart. PaGMIL trains a distinct classification head per task, which is linked to its prompts. For a new WSI, the appropriate classification head is selected based on its prompt for prediction.

\subsection{Patch Selector}

According to CLAM \cite{lu2021data}, we score and rank each patch to select positive and negative patches. Negative patches represent patches that are not important for the WSI and can be selected as $B$ patches with the lowest scores, denoted as $\mathcal{N}$. $B$ is a parameter that needs to be set, representing the number of patches. Positive patches, as representative patches of the WSI, need to be carefully selected. We first identify the top $k\%$ of patches with the highest scores. These patches are then stored as candidate positive patches, denoted as $\mathcal{P}_c$.

In cancer tissue slides, representative patches must contain cancerous cells, and cancer tissues always cluster together\cite{le2024sparsexmil}. Utilizing this knowledge, we can filter out some isolated high-scoring patches because these patches are usually staining bias or noise\cite{tellez2019quantifying}. We record each patch's position to calculate distances. For each patch in $\mathcal{P}_c$, if there is any other patch in $\mathcal{P}_c$ adjacent to it, both the current patch and its neighbors are added to $\mathcal{P}_s$.

Within the same WSI, cancer cells and their surrounding environments may vary across regions, which brings patch diversity. This diversity helps the model focus on different features, allowing the model to acquire generalizable knowledge that mitigates catastrophic forgetting. In order to find high-score patches that are far away as positive patches, we perform K-means clustering on the patches in $\mathcal{P}_s$ based on location, resulting in $B$ clusters. The center patches of these clusters serve as the $B$ positive patches.

We then feed the positive and negative patches into the instance classifier like in CLAM\cite{lu2021data} and supervise it using the smooth SVM Loss $\mathcal{L}_{SSVM}$ \cite{berrada2018smooth}.

% ==============================================
% ==============================================

\begin{table*}[t]
\caption{Quantitative comparison of different methods trained sequentially.}
\label{tab:mainexperiment}
\centering
\begin{threeparttable}
% \small 
% \scriptsize
% \scalebox{1.20}{ \setlength{\tabcolsep}{10pt}
%\setlength{\tabcolsep}{3pt} 
% \renewcommand{\arraystretch}{1} % 减小行间距

\begin{tabular}{l|cc|cc|cc|cc}
\bottomrule
\multirow{2}{*}{Method} & \multicolumn{2}{c|}{CAMELYON16 ($D_1$)} & \multicolumn{2}{c|}{BRACS ($D_2$)} & \multicolumn{2}{c|}{TCGA ($D_3$)} & \multicolumn{2}{c}{BACH ($D_4$)}\\
% \cline{2-7}
 & ACC & AUC & ACC & AUC & ACC & AUC & ACC & AUC  \\
 \hline
Separate   & 79.06 & 79.03 & 87.03 & 93.32 & 92.31 & 92.81 & 83.84 & 91.72 \\
\hline
CLAM\cite{lu2021data} (Baseline)  & 37.98 & 45.44 & 62.96 & 86.36 & 80.76 & 83.43 & 83.33 & 89.43 \\
EWC\cite{kirkpatrick2017overcoming} (PNAS2017)   & 37.98 & 58.39 & 68.51 & 90.19 & 84.23 & 85.12 & 85.25 & 90.49 \\
LWF\cite{li2017learning} (TPAMI2017)   & 37.98 & 47.73 & 76.62 & 85.36 & 86.54 & 92.50  & 84.44 & 92.35 \\
DER++\cite{buzzega2020dark} (NIPS2020)  & 38.75 & 47.60 & 81.48 & 89.48 & 84.61 & 90.62 & 76.56 & 86.70 \\
MIND \cite{bonato2024mind} (AAAI2024)  & 41.86 & 60.45 & 84.81 & 91.32 & 85.38 &  93.74 & 82.26 & 89.58 \\
PEC  \cite{zajkac2023prediction} (ICLR2024)   & 39.53 & 59.38 & 78.42 & 82.75 & 89.21 &  90.69 & 84.04 & 90.68 \\
ConSlide\cite{huang2023conslide} (ICCV2023)   & 48.53 & 59.45 & 81.85 & 88.42 & 88.46 & 91.62 & 81.61 & 91.33 \\
\bf PaGMIL (Ours)   & \bf 62.40 & \bf 68.64 & \bf 85.18 & \bf 92.13 & \bf 90.38 & \bf 93.74  & \bf 85.45 & \bf 92.85 \\
\bottomrule
\end{tabular}
\begin{tablenotes}
    
    \footnotesize
    \item[] \textbf{Note}: 
    Seperate denotes training a separate model for each dataset as a theoretical upper bound.
\end{tablenotes}
% }
\end{threeparttable}
\end{table*}

% ==============================================
% ==============================================

\subsection{Prompt Guide}

For each WSI, we stitch all tissue-containing patches into a large square and resize it to fit the Generator's input. We use a ResNet\cite{he2016deep} as the generator to convert the thumbnails into 768-dimensional features, which serve as the prompt. This approach can retain macroscopic information such as color to infer which dataset the WSI belongs to.

During the training process, we generate prompts for the different data with each task. Once the training for task $i$ is completed, we calculate the average value of the prompts and designate it as the prompt $m_i$ for that task. It is crucial to minimize the differences within the prompts of the same task and maximize the differences between the prompts of different tasks. Therefore, we need to design two different types of losses to control the generation of the prompts.

Intra-class loss, to bring prompts of the same task closer:
\begin{equation}
    \mathcal{L}_{intra} = \frac{1}{2} * \sum_{i=1}^{N}||m_i - \Bar{m}||^2,
\end{equation}
where $m_i$ represents the prompt generated for each WSI during the training process, and $\Bar{m}$ represents the average prompt generated in the current epoch. At the start of a new epoch, $\Bar{m}$ is recalculated. $||\dots||$ denotes the L2 (Euclidean) distance.

Inter-class loss, to keep prompts of different tasks apart:
\begin{equation}
    \mathcal{L}_{inter} = - \frac{1}{2NT} \sum_{t=1}^T \sum_{i=1}^{N} \left[ {d_{it}}^2 + \max(0, {min} - d_{it})^2 \right],
\end{equation}
\begin{equation}
    d_{it} = ||m_i - \Bar{m}_t||,
\end{equation}
where T represents the number of tasks trained previously, N represents the number of data points in the current task, $d_{it}$ represents the Euclidean distance between the current prompt and the previous average prompt, and $min$ is the minimum distance set, giving a greater penalty when the prompts of different tasks are closer to each other.

During testing, the thumbnail of the WSI is sent to the Prompt Generator, and the generated prompt is used to calculate the cosine similarity with the stored prompts. The classification head corresponding to the most similar prompt is selected. This typically implies that the style of the test data closely aligns with the style of the training set associated with a certain stored classification head.

\section{Experiment}
\subsection{Datasets}

We select breast cancer datasets from four distinct countries and regions. These datasets exhibit significant variations in staining and distribution. The datasets include: 1) \textit{CAMELYON16}, Netherlands challenge dataset with official train/test sets; 2) \textit{BRACS}, an Italian breast cancer dataset , from which we select 4 classes: normal, benign, in situ carcinoma, and invasive carcinoma; 3) \textit{TCGA-BRCA}, a public WSI database in the United States, from which we chose a subset of images with relatively consistent distribution; and 4) \textit{BACH}, a Portuguese breast cancer dataset, from which we chose the portion labeled as ``Photos``.

\subsection{Experimental settings}

We define a sequential order for the four datasets: \textit{CAMELYON16}, \textit{BRACS}, \textit{TCGA-BRCA}, and \textit{BACH}. This order places the most difficult dataset, CAMELYON16, at the beginning, puts the simple data in the middle, and places the relatively difficult data at the end. This arrangement poses the greatest challenge to the model's robustness. We evaluate the final performance of the model by measuring the Area Under the Curve (AUC) and accuracy (ACC).

We compare our method with the following approaches: 1) CLAM (Baseline)\cite{lu2021data}, sequentially training the model for newly arrived task; 2) EWC\cite{kirkpatrick2017overcoming}, a classic regularization-based method; 3) LWF\cite{li2017learning}, a classic knowledge distillation-based method; 4) DER++ \cite{buzzega2020dark}, a classic data replay-based method; 5) MIND\cite{bonato2024mind} and PEC\cite{zajkac2023prediction}, the state-of-the-art methods for regularization and distillation; 6) ConSlide\cite{huang2023conslide}, a method for continual learning on WSI.

\subsection{Implementation details}

We crop regions of 224$\times$224 pixels from the segmented foreground tissue at a 20x magnification to obtain patch-level images. Then we utilize Ctranspath\cite{wang2022transformer} to extract features from the corresponding images. The previously mentioned parameter $B$ is set to 8, meaning that PaGMIL will select 8 positive patches and 8 negative patches. Each task throughout experiment is conducted for 30 epochs. The experiments are run on an NVIDIA GeForce RTX 4090 GPU.
% ==============================================
% ==============================================

\begin{figure}[htbp]
    \centering
    \subfigure[Score heatmap of CLAM (Baseline) method]{\includegraphics[width=1.0\linewidth]{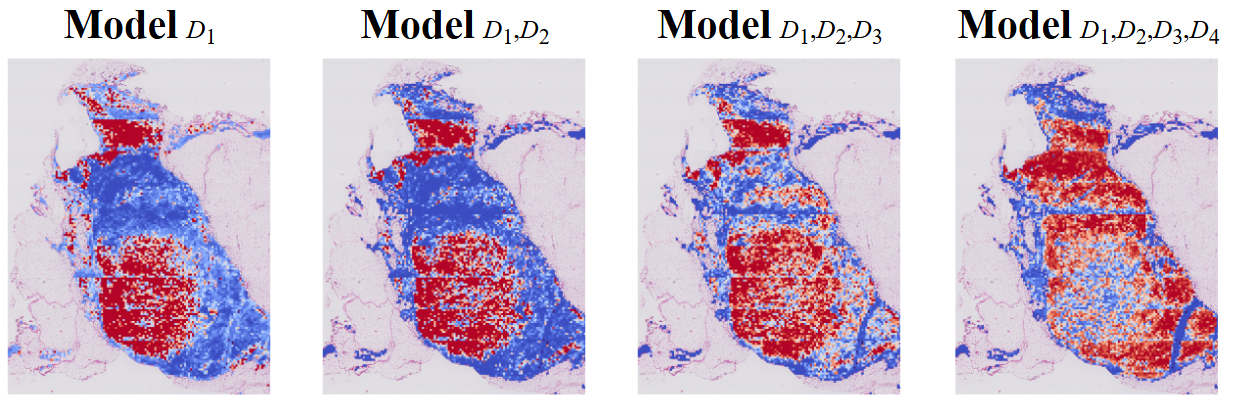}}
    \subfigure[Score heatmap of our method]{\includegraphics[width=1.0\linewidth]{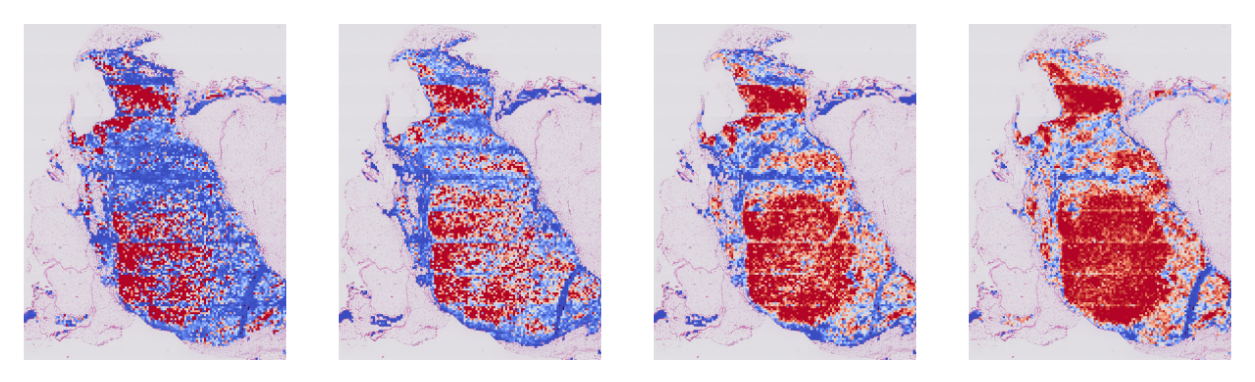}}
    
    \caption{The image is from CAMELYON16, which is WSI with cancer. Red indicates high scores given by the scoring network, while blue indicates low scores. The models are trained sequentially from left to right in the order of datasets. As training progresses, the baseline method shows an opposite distribution of red and blue, changing the representative patches from cancerous to normal tissue. In contrast, our method maintains a similar distribution of red and blue.}
    \label{fig:attention}
\end{figure}

% ==============================================
% ==============================================

\subsection{Main results}

Our method achieves good results in mitigating catastrophic forgetting according Table \ref{tab:mainexperiment}. It is worth noting that, following sequential training, the baseline, EWC\cite{kirkpatrick2017overcoming}, and LWF\cite{li2017learning} models all achieve a relatively low accuracy of 37.98\% when tested on the \textit{CAMELYON16} dataset. This is attributed to these models predicting all data as cancerous. We print out a score heatmap (see Fig.\ref{fig:attention}) of a WSI from \textit{CAMELYON16}, which might explain this phenomenon. 
The figure shows a heatmap based on the scoring, which is then weighted and used by the final classification head to determine whether the entire image is cancerous. In the Baseline method, the scoring network is always trained together with the classification head, leading them to make consistent decisions. After sequential training, the scoring network assigns high scores to normal tissue and low scores to cancerous tissue, misclassifying normal tissue as cancerous. Our method freezes the current classification head after training on one dataset, allowing the scoring network to focus solely on distribution of cancerous regions. Similar color distribution means that our PS module effectively captures generalizable knowledge and mitigates forgetting.

% ==============================================
% ==============================================

\begin{table}[ht]
\caption{Ablation Study on the Impact of Each Module.}
\label{tab:ablation1}
\centering
% \small 
% \scriptsize
% \setlength{\tabcolsep}{5pt}
% \setlength{\tabcolsep}{10pt} 
% \renewcommand{\arraystretch}{1} % 减小行间距
\resizebox{\columnwidth}{!}{
\begin{threeparttable}
\begin{tabular}{cc|cc|cc|cc|cc}
\bottomrule
\multirow{2}{*}{PS} & \multirow{2}{*}{PG} & \multicolumn{2}{c|}{CAMELYON16 ($D_1$)} & \multicolumn{2}{c|}{BRACS ($D_2$)} & \multicolumn{2}{c|}{TCGA ($D_3$)} & \multicolumn{2}{c}{BACH ($D_4$)}\\
% \cline{2-7}
& & ACC & AUC & ACC & AUC & ACC & AUC & ACC & AUC  \\
 \hline

 &  & 37.98 & 45.44 & 62.96 & 86.36 & 80.76 & 83.43 & 83.33 & 89.43 \\
$\checkmark$ &  & 39.53 & 52.80 & 69.25 & \bf 92.75 & 82.31 & 85.00  & 84.84 & 90.62 \\
 & $\checkmark$ & 58.29 & 64.81 & 82.96 & 89.94 & 88.85 & 92.69  & 85.05 & 92.58 \\
$\checkmark$ & $\checkmark$ &  \bf 62.40 & \bf 68.64 & \bf 85.18 & 92.13 & \bf 90.38 & \bf 93.74  & \bf 85.45 & \bf 92.85 \\

\bottomrule
\end{tabular}

\end{threeparttable}
}
\end{table}

% ==============================================
% ==============================================
\subsection{Ablation study}

To validate the effectiveness of the two components in our method, we conduct ablation experiments, and results are shown in Table \ref{tab:ablation1}. When using only the PS module, our results show improvement, indicating that the diversity of patches can help the model learn more generalizable knowledge to reduce catastrophic forgetting. Using only the PG module significantly improves results. This is because forgetting mainly occurs in the classification head. Using separate heads for each task and selecting the correct one during testing effectively reduces catastrophic forgetting.

\section{Conclusion}
In this paper, we propose PaGMIL, a novel architecture for mitigating catastrophic forgetting in breast cancer diagnosis. The PS module selects more accurate and diverse representative patches based on microscopic pathological prior knowledge, while the PG module determines the appropriate classification head for each WSI based on macroscopic pathological knowledge. Extensive experiments on four breast cancer datasets demonstrate that PaGMIL achieves a superior balance between current task effectiveness and retention of previous knowledge. In the future, we will conduct continual learning research on different types of cancer.

\bibliographystyle{IEEEtran}
\bibliography{main}

\end{document}